\documentstyle[aclap,lingmacros]{article}

\newcommand{\secref}[1]{Section~\ref{#1}}
\newcommand{\figref}[1]{Figure~\ref{#1}}

\newcommand{\veps}{\varepsilon}

\newcommand{\uvd}{{\rm UVG-DL}}

\newcommand{\size}[1]{\mbox{$\mid \! {#1} \! \mid$}}

\newcommand{\olproof}{\noindent{\bf Proof (outline).\hspace{1em}}}
\newcommand{\closeproof}{\mbox{\hspace{1em}\rule{.45em}{.45em}}}

\newcommand{\deri}{\mbox{$\:\stackrel{\!\!\!\!*}{\Longrightarrow}\:$}}

\newcommand{\vdm}[1]{\mbox{${\vdash_{\!\!\!\!{\mit {\scriptscriptstyle
M}}}}\:$}}
\newcommand{\vdsm}[1]{\mbox{$\:\:\:\raisebox{.45em}{{$\scriptstyle *$}}\hspace{-.62em}{\vdash_{\!\!\!\!{\mit {\scriptscriptstyle M}}}}\:$}}

\newtheorem{theorem}{Theorem}
\newtheorem{lemma}[theorem]{Lemma}

\newcounter{exacount}
\newcommand{\uvgd}{\mbox{{\rm UVG-DL}}}
\newcommand{\suvgd}{\mbox{{\rm SynchUVG-DL}}}
\newcommand{\suvd}{\mbox{{\rm SynchUVG-DL}}}
\newcommand{\nlstag}{\mbox{{\rm nlSynchTAG}}}
\newcommand{\stag}{\mbox{{\rm SynchTAG}}}

\newcommand{\out}{\mbox{${\it out}$}}

\include{psfig}

\newcommand{\namecite}[1]{\cite{#1}}

\title{\vspace{-1in}{\small In {\em Proceedings of the 34th Meeting of the Association for Computational Linguistics (ACL'96)}}\\
Synchronous Models of Language}

\author{Owen Rambow \\
CoGenTex, Inc.                       \\
840 Hanshaw Road, Suite 11           \\
Ithaca, NY 14850-1589                \\
{\tt owen@cogentex.com}\And
       Giorgio Satta \\
        	Dipartimento di Elettronica ed Informatica \\ 
        	Universit\`{a} di Padova \\
        	via Gradenigo, 6/A \\
        	I-35131 Padova,	Italy \\
	{\tt satta@dei.unipd.it}}

\begin{document}

\maketitle
\vspace{-0.5in}
\begin{abstract}
In synchronous rewriting, the productions of two rewriting systems are
paired and applied synchronously in the derivation of a pair of strings.
We present a new synchronous rewriting system and argue that it can handle
certain phenomena that are not covered by existing synchronous systems.  We
also prove some interesting formal/computational properties of our system.
\end{abstract}

\section{Introduction}

Much of theoretical linguistics can be formulated in a very natural manner
as stating correspondences (translations) between layers of representation;
for example, related interface layers LF and PF in GB and Minimalism
\cite{chomsky:1993}, semantic and syntactic information in HPSG
\cite{pollard/sag:1994}, or the different structures such as c-structure
and f-structure in LFG \cite{bresnan/kaplan:1982}.  Similarly, many
problems in natural language processing, in particular parsing and
generation, can be expressed as transductions, which are calculations of
such correspondences.  There is therefore a great need for formal models of
corresponding levels of representation, and for corresponding algorithms
for transduction.

Several different transduction systems have been used in the past by the
computational and theoretical linguistics communities.  These systems have
been borrowed from translation theory, a subfield of formal language
theory, or have been originally (and sometimes redundantly) developed.
{\em Finite state transducers} (for an overview, see, e.g.,
\cite{aho/ullman:1972}) provide translations between regular languages.
These devices have been popular in computational morphology and
computational phonology since the early eighties
\cite{koskenniemi:1983,kaplan/kay:1994}, and more recently in parsing as
well (see, e.g., \cite{gross:1989a,pereira:1991,roche:1993}).  {\em
Pushdown transducers} and {\em syntax directed translation schemata} (SDTS)
\cite{aho/ullman:1969a} translate between context-free languages and are
therefore more powerful than finite state transducers.  Pushdown
transducers are a standard model for parsing, and have also been used
(usually implicitly) in speech understanding.  Recently, variants of SDTS
have been proposed as models for simultaneously bracketing parallel
corpora~\cite{wu:1995}.  Synchronization of tree adjoining grammars (TAGs)
\cite{shieber/schabes:1990a,shieber:1994} are even more powerful than the
previous formalisms, and have been applied in machine translation
\cite{ASJ:1990,egedi/palmer:1994,harbusch/poller:1994,prigent:1994},
natural language generation \cite{shieber/schabes:1991}, and theoretical
syntax \cite{abeille:1994a}.  The common underlying idea in all of these
formalisms is to combine two generative devices through a pairing of their
productions (or, in the case of the corresponding automata, of their
transitions) in such a way that right-hand side nonterminal symbols in the
paired productions are {\em linked}.  The processes of derivation proceed
synchronously in the two devices by applying the paired grammar rules only
to linked nonterminals introduced previously in the derivation.  The fact
that the above systems all reflect the same translation technique has not
always been recognized in the computational linguistics literature.
Following~\namecite{shieber/schabes:1990a} we will refer to the general
approach as {\em synchronous rewriting}.  While synchronous systems are
becoming more and more popular, surprisingly little is known about the
formal characteristics of these systems (with the exception of the
finite-state devices).

In this paper, we argue that existing synchronous systems cannot handle, in
a computationally attractive way, a standard problem in syntax/semantics
translation, namely quantifier scoping.  We propose a new system that
provides a synchronization between two unordered vector grammars with
dominance links (\uvd) \namecite{rambow:1994c}.  The type of
synchronization is closely based on a previously proposed model, which we
will call ``local'' synchronization.  We argue that this synchronous system
can deal with quantifier scoping in the desired way.  The proposed system
has the {\em weak language preservation property}, that is, the defined
synchronization mechanism does not alter the weak generative capacity of
the formalism being synchronized.  Furthermore, the {\em tree-to-forest
translation\/} problem for our system can be solved in polynomial time;
that is, given a derivation tree obtained according to one of the
synchronized grammars, we can construct the forest of all the translated
derivation trees in the other grammar, using a polynomial amount of time.

The structure of this paper is as follows.  In \secref{sec-qr}, we
introduce quantifier raising and review two types of synchronization and
mention some new formal results.  We introduce our new synchronous system
in~\secref{sec-def}, and present our formal results and outline the proof
techniques in \secref{sec-proofs}.

\section{Types of Synchronization}

\label{sec-qr}

\subsection{Quantifier Raising}

We start by presenting an example which is based on transfer between a
syntactic representation and a ``semantic'' representation of the scoping
of quantified NPs.  It is generally assumed that in English (and many other
languages), quantified arguments of a verb can (in appropriate contexts)
take scope in any possible order, and that this generalization extends to
cases of embedded clauses \cite{may:1985}.\footnote{We explicitly exclude
from our analysis cases of quantified NPs embedded in NPs, and do not, of
course, propose to develop a serious linguistic theory of quantifier
scoping.}  For example, sentence (\ref{sent-exa}) can have four possible
interpretations (of the six possible orderings of the quantifiers, two
pairs are logically equivalent), two of which are shown in
(\ref{sec-scopes}).

\enumsentence{\label{sec-exa}\label{sent-exa}
Every man thinks some official said some Norwegian arrived}

\eenumsentence{\label{sec-scopes}\label{sent-scopes}
\item
$\forall x$, $x$ a man, $\exists y$, $y$ an official, $\exists z$, $z$ a
Norwegian, $x$ thinks $y$ said $z$ arrived
\item
$\exists z$, $z$ a Norwegian, $\exists y$, $y$ an official, $\forall x$,
$x$ a man, $x$ thinks $y$ said $z$ arrived }

\begin{figure}[htb]
\centerline{
\psfig{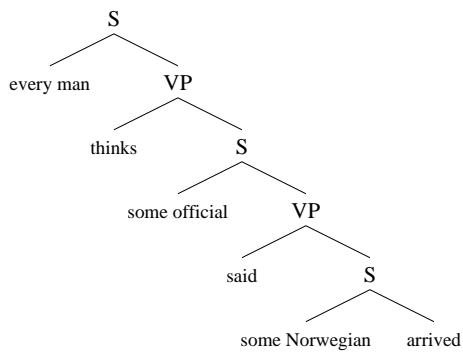}
}
\caption{Syntactic representation for (1)}
\label{fig-exa}\label{fig-syntax}
\end{figure}

\begin{figure}[htb]
\centerline{
\psfig{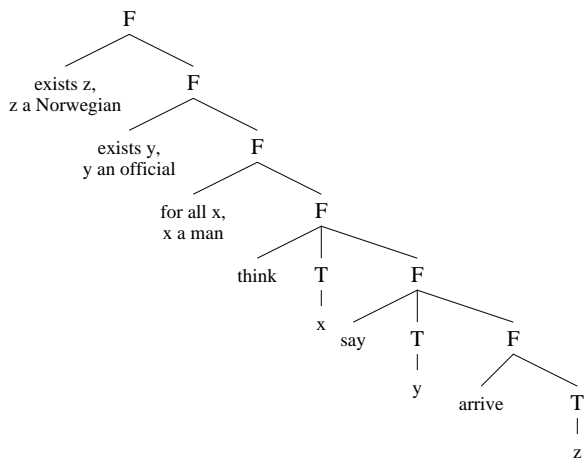}
}
\caption{Semantic representation for (2b)}
\label{fig-scopes}\label{fig-semantics}
\end{figure}

We give a simplified syntactic representation for (\ref{sent-exa}) in
\figref{fig-exa}, and a simplified semantic representation for
(\ref{sec-scopes}b) in \figref{fig-scopes}.

\subsection{Non-Local Synchronization}

We will first discuss a type of synchronization proposed by
\namecite{shieber/schabes:1990a}, based on TAG.  We will refer to this
system as {\em non-local synchronous TAG} (\nlstag).  The synchronization
is non-local in the sense that once links are introduced during a
derivation by a synchronized pair of grammar rules, they need not continue
to impinge on the nodes that introduced them: the links may be reassigned
to a newly introduced nonterminal when an original node is rewritten.  We
will refer to this mechanism as {\em link inheritance}.  To illustrate, we
will give as an example an analysis of the quantifier-raising example
introduced above, extending in a natural manner an example given by Shieber
and Schabes.

\begin{figure}[htb]
\centerline{
\psfig{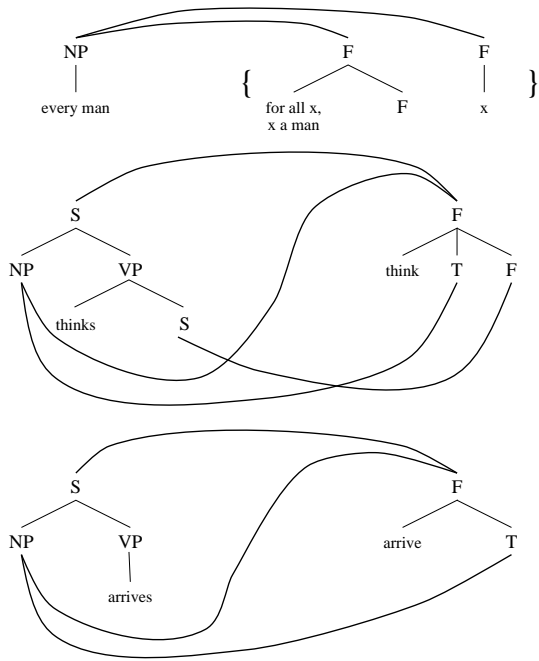}
}
\caption{Elementary structures in nlSynchTAG}
\label{fig-nl-elems}
\end{figure}

\begin{figure}[tb]
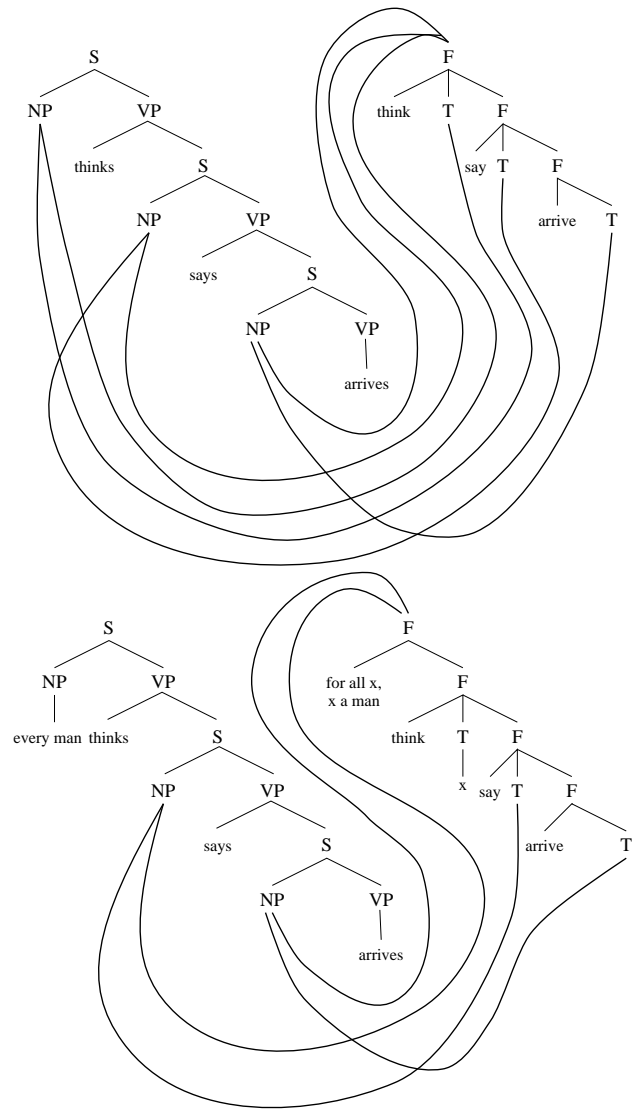

\centerline{
\psfig{figure=nlstag-der2-small.eps,scale=57}}
\centerline{
\psfig{figure=nlstag-der3-small.eps,scale=57}}
\caption{Non-local derivation  in nlSynchTAG}
\label{fig-nl-der1}
\end{figure}

The elementary structures are shown in \figref{fig-nl-elems} (we only give
one NP --- the others are similar).  The nominal arguments in the syntax
are associated with pairs of trees in the semantics, and are linked to two
nodes, the quantifier and the variable.  The derivation proceeds as
illustrated in Figure~\ref{fig-nl-der1}, finally yielding the two
structures in \figref{fig-syntax} and \figref{fig-scopes}.  Note that some
of the links originating with the NP nodes are inherited during the
derivation.  By changing the order in which we add the nominal arguments at
the end of the derivation, we can obtain all quantifier scopes in the
semantics.

The problem with non-local synchronization is that the weak language
preservation property does not hold.  \namecite{shieber:1994} shows that
not all {\nlstag} left-projection languages can be generated by TAGs.  As a
new result, in~\cite{rambow/satta:1996a} we show that the recognition of
some fixed left-projection languages of a {\nlstag} is NP-complete.  Our
reduction crucially relies on link inheritance.  This makes {\nlstag}
unattractive for applications in theoretical or computational linguistics.

\subsection{Local Synchronous Systems}

In contrast with non-local synchronization, in local synchronization there
is no inheritance of synchronization links.  This is enforced by requiring
that the links establish a bijection between nonterminals in the two
synchronously derived sentential forms, that is, each nonterminal must be
involved in exactly one link.  In this way, once a nonterminal is rewritten
through the application of a pair of rules to two linked nonterminals, no
additional link remains to be transferred to the newly introduced
nonterminals.  As a consequence of this, the derivation structures in the
left and right grammars are always isomorphic (up to ordering and labeling
of nodes).

\begin{figure}[htb]
\parbox{\columnwidth}{
{\bf Grammar}:\newline
\begin{small}
\begin{tabular}{ll}
S\framebox{{\tiny 1}} $\rightarrow$ NP\framebox{\tiny 2}
{\em likes} NP\framebox{\tiny 3} & S\framebox{\tiny 1}
$\rightarrow$ NP\framebox{\tiny 3} {\em pla{\^\i}t {\`a}}
NP\framebox{\tiny 2}\\
NP\framebox{\tiny  4} $\rightarrow$ {\em John} & NP\framebox{\tiny  4} $\rightarrow$ {\em Jean} \\
NP\framebox{\tiny  5} $\rightarrow$ {\em the white} N\framebox{\tiny  6} & NP\framebox{\tiny  5} $\rightarrow$ {\em la} N\framebox{\tiny  6} {\em
blanche} \\
N\framebox{\tiny 7} $\rightarrow$ {\em house} & N\framebox{\tiny 7} $\rightarrow$ {\em maison}
\end{tabular}
\end{small}
{\bf Derivation}:\newline
\begin{small}
(S\framebox{\tiny 0}, S\framebox{\tiny 0}) \\
\rule{.3cm}{0cm}$\Longrightarrow$(NP\framebox{\tiny 2} {\em likes} NP\framebox{\tiny 3}, NP\framebox{\tiny 3} {\em pla{\^\i}t {\`a}} NP\framebox{\tiny 2})\\
\rule{.3cm}{0cm}$\Longrightarrow$(NP\framebox{\tiny 2} {\em likes the white} N\framebox{\tiny 6}, {\em la} N\framebox{\tiny 6} {\em blanche pla{\^\i}t {\`a}} NP\framebox{\tiny 2})\\ 
\rule{.3cm}{0cm}$\deri$({\em John likes the white house}, {\em \rule{3mm}{0mm}la maison blanche pla{\^\i}t {\`a}
Jean}) 
\end{small}
}
\caption{Sample SDTS and derivation}
\label{fig-sdts}
\end{figure}

The canonical example of local synchronization is SDTS
\cite{aho/ullman:1969a}, in which two context-free grammars are
synchronized.  We give an example of an SDTS and a derivation in
\figref{fig-sdts}.  The links are indicated as boxed numbers to the right
of the nonterminal to which they apply.  \cite{shieber:1994} defines the
tree-rewriting version of SDTS, which we will call {\em synchronous TAG}
(\stag), and argues that {\stag} does not have the formal problems of
{\nlstag} (though \cite{shieber:1994} studies the translation problem
making the unappealing assumption that each tree in the input grammar is
associated with only one output grammar tree).

However, {\stag} cannot derive all possible scope orderings, because of the
locality restriction.  This can be shown by adapting the proof technique in
\cite{BRN:1992-short}.  In the following section, we will present a
synchronous system which has local synchronization's formal advantages, but
handles the scoping data.

\section{Extended Local Synchronization}

\label{sec-def}

\begin{figure}[htb]
\centerline{
\psfig{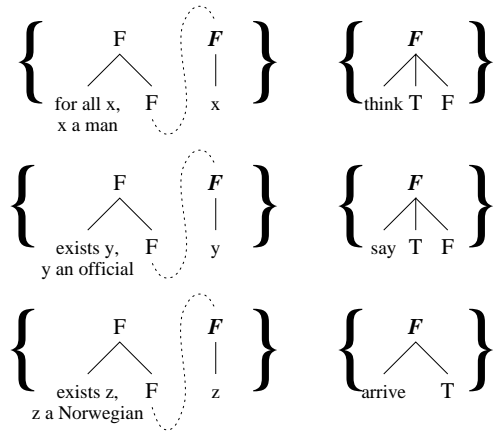}}
\caption{A UVG-DL for deriving semantic representations such as (2)}
\label{fig-uvd-gram}
\end{figure}

\begin{figure}[htb]
\centerline{
\psfig{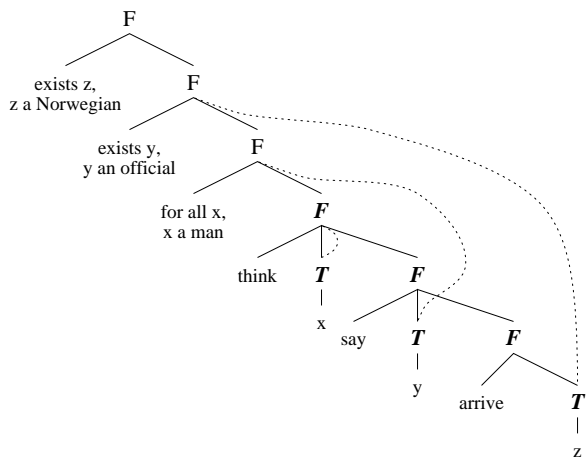}
}
\caption{Derivation of (2b) in a UVG-DL}
\label{fig-uvd-der}
\end{figure}

In this section, we propose a new synchronous system, which is based on
local synchronization of unordered vector grammars with dominance links
(\uvd)~\cite{rambow:1994c}.  The presentations will be informal for reasons
of space; we refer to~\cite{rambow/satta:1996a} for details.  In \uvd ,
several context-free string rewriting rules are grouped into sets, called
{\em vectors}.  In a derivation, all or no rules from a given instance of a
vector must be used.  Put differently, all productions from a given vector
must be used the same number of times.  They can be applied in any order
and need not be applied simultaneously or one right after the other.  In
addition, \uvd\/ has {\em dominance links}.  An occurrence of a nonterminal
$A$ in the right-hand side of a rule $p$ can be linked to the left-hand
nonterminal of another rule $p'$ in the same vector.  This dominance link
will act as a constraint on derivations: if $p$ is used in a derivation,
then $p'$ must be used subsequently in the subderivation that starts with
the occurrence of $A$ introduced by $p$.  A {\uvd} is {\em lexicalized} iff
at least one production in every vector contains a terminal symbol.
Henceforth, all \uvd s mentioned in this paper will implicitly be assumed
to be lexicalized.  The derivation structure of a {\uvd} is just the
derivation structure of the same derivation in the underlying context-free
grammar (the CFG obtained by forming the union of all vectors).  We give an
example of a {\uvd} in \figref{fig-uvd-gram}, in which the dotted lines
represent the dominance links.  A sample derivation is in
\figref{fig-uvd-der}.

Our proposal for the synchronization of two UVG-DL uses the notion of
locality in synchronization, but with respect to entire vectors, not
individual productions in these vectors.  This approach, as we will see,
gives us both the desired empirical coverage and acceptable computational
and formal results.  We suppose that in each vector $v$ of a {\uvd} there
is exactly one privileged element, which we call the {\em synchronous
production} of $v$.  All other elements of $v$ are referred to as {\em
asynchronous productions}.  In Figures~\ref{fig-uvd-gram}
and~\ref{fig-uvd-der}, the synchronous productions are designated by a
bold-italic left-hand side symbol.  Furthermore, in the right-hand side of
each asynchronous production of $v$ we identify a single nonterminal,
called the {\em heir}.

\begin{figure}[tb]
\centerline{
\psfig{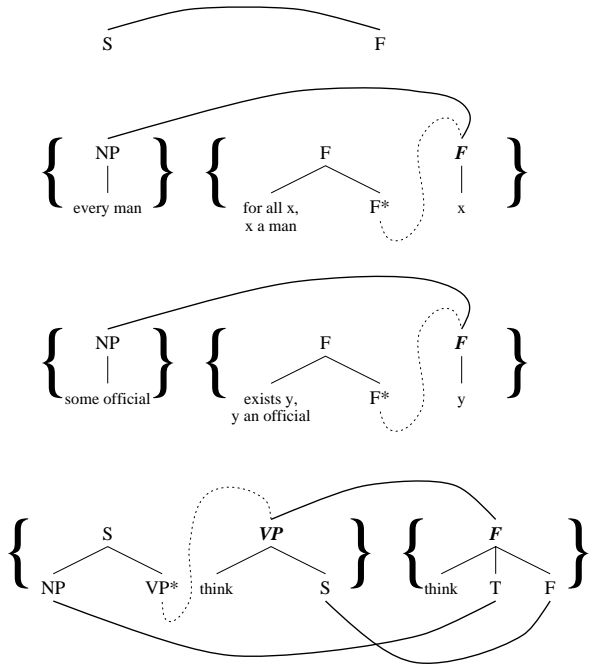}
}
\caption{{\suvd} grammar for quantifier scope disambiguation}
\label{fig-suvd-gram}
\end{figure}

In a {\em synchronous UVG-DL} (SynchUVG-DL), vectors from one \uvd\/ are
synchronized with vectors from another \uvd .  Two vectors are synchronized
by specifying a bijective synchronization mapping (as in local
synchronization) between the non-heir right-hand side occurrences of
nonterminals in the productions of the two vectors.  A nonterminal on which
a synchronization link impinges is referred to as a {\em synchronous
nonterminal}.  A sample {\suvd} grammar is shown in \figref{fig-suvd-gram}.

Informally speaking, during a SynchUVG-DL derivation, the two synchronous
productions in a pair of synchronized vectors must be applied at the same
time and must rewrite linked occurrences of nonterminals previously
introduced.  The asynchronous productions of the two synchronized grammars
are not subject to the synchronization requirement, and they can be applied
at any time and independently of the other grammar (but of course subject
to the grammar-specific dominance links).  Any synchronous links that
impinge on a nonterminal rewritten by an asynchronous production are
transferred to the heir of the asynchronous production.  (A production may
introduce a synchronous nonterminal whose counterpart in the other grammar
has not yet been introduced.  In this case, the link remains ``pending''.)
Thus, while in SynchUVG-DL there is link inheritance as in non-local
synchronization, link inheritance is only possible with those productions
that themselves are not subject to the synchronization requirement.

\begin{figure}[htb]
\centerline{
\psfig{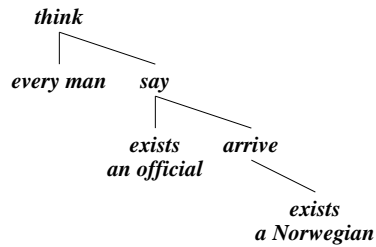}
}
\caption{Vector derivation tree for derivation of (2b)}
\label{fig-uvd-vecder}
\end{figure}

The locality of synchronization in SynchUVG-DL becomes clear when we
consider a new structure which we introduce here, called the {\em vector
derivation tree}.  Consider two synchronized \uvd derivations in a
SynchUVG-DL.  The vector derivation tree for either component derivation is
obtained as follows.  Each instance of a vector used in the derivation is
represented as a single node (which we label with that vector's lexeme).  A
node representing a vector $v_1$ is immediately dominated by the node
representing the vector $v_2$ which introduced the synchronization link
that the synchronous production of $v_1$ rewrites.  Unlike the standard
derivation tree for \uvd , the vector derivation tree clearly shows how the
vectors (rather than the component rules of the vectors) were combined
during the derivation.  The vector derivation tree for the derivation in
\figref{fig-uvd-der} is shown in \figref{fig-uvd-vecder}.

It should be clear that the vector derivation trees for two
synchronized derivations are isomorphic, reflecting the fact that our
definition of SynchUVG-DL is local with respect to vectors (though not
with respect to productions, since the derivation trees of two
synchronized UVG-DL derivations need not be isomorphic).  The vector
derivation tree can be seen as representing an ``outline'' for the
derivation.  Such a view is attractive from a linguistic perspective:
if each vector represents a lexeme and its projection (where the
synchronous production is the basis of the lexical projection that the
vector represents), then the vector derivation tree is in fact the
dependency tree of the sentence (representing direct relations between
lexemes such as grammatical function).  In this respect, the vector
derivation tree of \uvd\/ is like the derivation tree of tree
adjoining grammar and of D-tree grammars (DTG)~\cite{RVW:1995a}, which is
not surprising, since all three formalisms share the same extended domain
of locality.  Furthermore, the vector derivation tree of SynchUVG-DL shares
with the the derivation tree of DTG the property that it reflects
linguistic dependency uniformly; however, while the definition of DTG was
motivated precisely from considerations of dependency, the vector
derivation tree is merely a by-product of our definition of SynchUVG-DL,
which was motivated from the desire to have a computationally tractable
model of synchronization more powerful than SynchTAG.%
\footnote{We do not discuss modifiers in this
paper for lack of space.}

\begin{figure}[htb]
\centerline{
\psfig{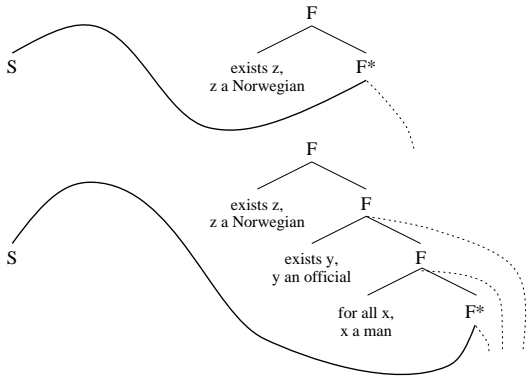}
}
\caption{{\suvd} derivation, steps 1 and 2}
\label{fig-suvd-der2}
\end{figure}

\begin{figure}[hbt]
\centerline{
\psfig{figure=suvd-der3-small.eps,scale=57}
}
\caption{{\suvd} derivation, step 3}
\label{fig-suvd-der3}
\end{figure}

We briefly discuss a sample derivation.  We start with the two start
symbols, which are linked.  We then apply an asynchronous production from
the semantic grammar.  In \figref{fig-suvd-der2} (top) we see how the link is
inherited by the heir nonterminal of the applied production.  This step is
repeated with two more asynchronous productions, yielding
\figref{fig-suvd-der2} (bottom).  We now apply productions for the bodies of the
clauses, but stop short before the two synchronous productions for the {\em
arrive} clause, yielding \figref{fig-suvd-der3}.  We see the asynchronous
production of the syntactic {\em arrive} vector has not only inherited the
link to its heir nonterminal, but has introduced a link of its own.  Since
the semantic end of the link has not been introduced yet, the links remains
pending.  We then finish the derivation to obtain the
two trees in \figref{fig-syntax} and \figref{fig-semantics}, with no
synchronization or dominance links left.

\section{Formal results}

\label{sec-proofs}

\begin{theorem}
$\suvgd$ has the language preservation property. 
\end{theorem}

\olproof Let $G_s$ be a $\suvgd$, $G'$ and $G''$ its left and right $\uvgd$
components, respectively.  We construct a $\uvgd$ $G$ generating the
left-projection language of $G_s$.  $G$ uses all the nonterminal symbols of
$G'$ and $G''$, and some compound nonterminals of the form $[A,B]$, $A$ and
$B$ nonterminals of $G'$ and $G''$, respectively.  $G$ simulates $G_s$
derivations by intermixing symbols of $G'$ and symbols of $G''$, and
without generating any of the terminal symbols of $G''$.  Most important,
each pair of linked nonterminals generated by $G_s$ is represented by $G$
using a compound symbol.  This enforces the requirement of simultaneous
application of synchronous productions to linked nonterminals.

Each vector $v$ of $G$ is constructed from a pair of synchronous vectors
$(v',v'')$ of $G_s$ as follows.  First, all instances of nonterminals in
$v''$ are replaced by $\veps$.  Furthermore, for any instance $B$ of a
right-hand side nonterminal of $v''$ linked to a right-hand side
nonterminal $A$ of $v'$, $B$ is replaced by $\veps$ and $A$ by $[A,B]$.
Then the two synchronous productions in $v'$ and $v''$ are composed into a
single production in $v$, by composing the two left-hand sides in a
compound symbol and by concatenating the two right-hand sides.  Finally, to
simulate link inheritance in derivations of $G_s$, each asynchronous
production in $v'$ and $v''$ is transferred to $v$, either without any
change, or by composing with some nonterminal $C$ both its left-hand side
and the heir nonterminal in its right-hand side.  Note that there are
finitely many choices for the last step, and each choice gives a different
vector in $G$, simulating the application of $v'$ and $v''$ to a set of
(occurrences of) nonterminals in a particular link configuration in a
sentential form of $G_s$.~\closeproof

We now introduce a representation for sets of derivation trees in a $\uvgd$
$G$.  A {\em parse tree\/} in $G$ is an ordered tree representing a
derivation in $G$ and encoding at each node the production $p$ used to
start the corresponding subderivation and the multiset of productions $f$
used in that subderivation.  A {\em parse forest\/} in $G$ is a directed
acyclic graph which is ordered and bipartite.  (We use ideas originally
developed in~\cite{lang:1991} for the context-free case.)  Nodes of the
graph are of two different types, called {\em and-nodes\/} and {\em
or-nodes}, respectively, and each directed arc connects nodes of different
types.  A parse forest in $G$ represents a set $T$ of parse trees in $G$ if
the following holds.  When starting at a root node and walking through the
graph, if we follow exactly one of the outgoing arcs at each or-node, and
all of the outgoing arcs at each and-node, we obtain a tree in $T$ modulo
the removal of the or-nodes.
Furthermore, every tree in $T$ can be obtained in this way.

\begin{lemma}
Let $G$ be a $\uvgd$ and let $q \geq 1$ be a natural number. 
The parse forest representing the set 
of all parse trees in $G$ with no more than 
$q$ vectors can be constructed in an amount of time 
bounded by a polynomial function of $q$.~\closeproof
\label{l:qforest}
\end{lemma}

Let $G_s$ be a $\suvgd$, $G'$ and $G''$ its left and
right $\uvgd$ components, respectively. 
For a parse tree $\tau$ in $G'$, we denote as 
$T(\tau)$ the set of all parse trees in $G''$ 
that are synchronous with $\tau$ according to $G_s$. 
The {\em parse-to-forest translation\/} problem for $G_s$ 
takes as input a parse tree $\tau$ in $G'$ and gives as output 
a parse forest representation for $T(\tau)$. 
If $G_s$ is lexicalized, such a parse forest has size bounded 
by a polynomial function of $\size{\tau}$, despite the fact that 
the size of $T(\tau)$ can be exponentially larger than the size
of $\tau$.  In fact, we have a stronger result. 

\begin{theorem}
The parse-to-forest translation problem for a lexicalized $\/\suvgd$ 
can be computed in polynomial time. 
\label{t:trans} 
\end{theorem} 

\olproof Let $G_s$ be a $\suvgd$ with $G'$ and $G''$ its left and right
$\uvgd$ components, respectively.  Let $\tau$ be a parse tree in $G'$ and
$\pi$ be the parse forest representing $T(\tau)$.  The construction of
$\pi$ consists of two stages.

In the first stage, we construct the vector derivation tree $\gamma$
associated with~$\tau$.  Let $q$ be the number of nodes of $\gamma$.  We
also construct a parse forest $\pi_q$ representing the set of all parse
trees in $G''$ with no more than $q$ vectors.  This stage takes polynomial
time in the size of $\tau$, since $\gamma$ can be constructed from $\tau$
in linear time and $\pi_q$ can be constructed as in Lemma~\ref{l:qforest}.

In the second stage, we remove from $\pi_q$ all the parse trees not in
$\pi$. This completes the construction, since the set of parse trees
represented by $\pi$ is included in the set of parse trees represented by
$\pi_q$.  Let $n_r$ and $\Gamma$ be the root node and the set of all nodes
of $\gamma$, respectively.  For $n \in \Gamma$, $\out(n)$ denotes the set
of all children of $n$.  We call {\em family\/} the set $\{n_r\}$ and any
nonempty subset of $\out(n)$, $n \in \Gamma$.  The main idea is to
associate a set of families ${\cal F}_n$ to each node $n$ of $\pi_q$, such
that the following condition is satisfied.  A family $F$ belongs to ${\cal
F}_n$ if and only if at least one subderivation in $G''$ represented at $n$
induces a forest of vector derivation trees whose root nodes are all and
only the nodes in $F$.  Each ${\cal F}_n$ can easily be computed visiting
$\pi_q$ in a bottom-up fashion.  Crucially, we ``block'' a node of $\pi_q$
if we fail in the construction of ${\cal F}_n$.  We claim that each set
${\cal F}_n$ has size bounded by the number of nodes in $\gamma$. This can
be shown using the fact that all derivation trees represented at a node of
$\pi_q$ employ the same multiset of productions of $G''$.  From the above
claim, it follows that $\pi_q$ can be processed in time polynomial in the
size of $\tau$.  Finally, we obtain $\pi$ simply by removing from $\pi_q$
all nodes that have been blocked.~\closeproof

\section{Conclusion}

We have presented {\suvd}, a synchronous system which has restricted
formal power, is computationally tractable, and which handles the
quantifier-raising data.  In addition, {\suvd} can be used for modeling the
syntax of languages with syntactic constructions which have been argued to
be beyond the formal power of TAG, such as scrambling in German and many
other languages \cite{rambow:1994c} or {\em wh}-movement in Kashmiri
\cite{RVW:1995a}.  {\suvd} can be used to synchronize a syntactic grammar
for these languages either with a semantic grammar, or with the syntactic
grammar of another language for machine translation applications.  However,
{\suvd} cannot handle the list of cases listed in \cite{shieber:1994}.
These pose a problem for {\suvd} for the same reason that they pose a
problem for other local synchronous systems: the (syntactic) dependency
structures represented by the two derivations are different.  These cases
remain an open research issue.

\section*{Acknowledgments}

Parts of the present research were done while Rambow was supported by the
North Atlantic Treaty Organization under a Grant awarded in 1993, while at
TALANA, Universit{\'e} Paris 7, and while Satta was visiting the Center for
Language and Speech Processing, Johns Hopkins University, Baltimore, MD.

\end{document}